\documentclass[11pt, a4paper]{article}
\usepackage[utf8]{inputenc}
\usepackage{graphicx}
\usepackage{latexsym}
\usepackage{amsmath,amssymb,cite,geometry,url}
\usepackage{mathptmx}

\newtheorem{theorem}{Theorem}
\newtheorem{remark}{Remark}

\newtheorem{lemma}{Lemma}

\DeclareMathOperator{\R}{\mathbb{R}}

\begin{document}
	\begin{center}
		{\large\bf{Deep Learning with Nonsmooth Objectives}}
	\end{center}
	
	\begin{center}
		{Vinesha Peiris, Vera Roshchina, Nadezda Sukhorukova}
	\end{center}

%maketitle

\begin{abstract}

We explore the potential for using a nonsmooth loss function based on the max-norm in the training of an artificial neural network. We hypothesise that this may lead to superior classification results in some special cases where the training data is either very small or unbalanced.

Our numerical experiments performed on a simple artificial neural network with no hidden layers (a setting immediately amenable to standard nonsmooth optimisation techniques) appear to confirm our hypothesis that uniform approximation based approaches may be more suitable for the datasets with reliable training data that either is limited size or biased in terms of relative cluster sizes. 

{\bf keywords}:
	quasiconvex functions, bisection method for quasiconvex minimisation, deep learning
	
\end{abstract}

%90C26, % Non-convex programming and global optimization
%90C90,                % Applications of mathematical programming
%90C47,              %minimax
%65D15,   % Algorithms for functional approximation
%65K10.

%\tableofcontents

\section{Introduction}

Deep learning is a popular tool in the modern area of Artificial Intelligence. Deep learning has many practical applications, including data analysis, signal and image processing and many others~\cite{Goodfellow2016,Sun2020OptimDeepLearning,MNIST}. %\todo{Ref \cite{MNIST} appears to be off}-fixed. 
Deep learning is based on solid mathematical modelling established in~\cite{Cybenko,Hornik,LeshnoPinkus1993,pinkus_1999}. These works demonstrate that deep learning solves approximation problems and, in its essence, relies on the results of the celebrated Kolmogorov-Arnold Theorem~\cite{Kol57,Arnold57}. There is a massive amount of publications and internet resources dedicated to deep learning. One of the most comprehensive and thorough textbook on the modern view on deep learning can be found in~\cite{Goodfellow2016}.  This book also touches upon the optimisation side of the problem. In particular, the goal is to minimise the loss function, which is also the measure of ``inaccuracy''. Overall, the goal of deep learning is to optimise weights in the network and therefore, this problem can be treated using modern optimisation tools~\cite{Sun2020OptimDeepLearning, VidalHaeffele17, Goodfellow2016,DeepDeclarative}.

It is customary to choose the mean least squares loss function to evaluate and optimise the performance of a neural network against a training set. There are several reasons for the popularity of the least squares formulation. From the optimisation perspective this model involves minimising a smooth quadratic objective function, and therefore basic optimisation techniques such as the gradient descent can be successfully employed. The least squares formulation also fares well with the assumption that the errors in the training set have a normal distribution. 

The goal of this work is to explore an alternative choice of loss functions and to analyse the impact these choices have on the quality of the training and the choice of the optimisation technique. The idea to use a different objective is not new, and was explored in the literature and applications. For instance, $l_1$-norm may be useful in the settings when sparsity is sought after \cite{sparse}. Other sources where alternative loss functions are explore are~\cite{Marcotte92,Goodfellow2016}.

%\todo[inline]{We need to do a bit more work with the literature review}

Our main hypothesis is that uniform approximation-based models may work better than least squares-based models whenever the size of the data available for training the model is small while highly reliable or when the data is unbalanced. This is the case for the kind of datasets where each observation is a results of a very expensive procedure or experiment and therefore the availability of data is limited~\cite{ExpensiveExperiments}. At the same time, the reliability of such data is high, since every experiment is carefully designed and analysed. Our preliminary numerical experiments appear to confirm the claim.

We use an elementary model of an artificial neural network that has no hidden layers and only one output node. It was fairly easy to implement the necessary nonsmooth optimisation technique for this model to test our assumptions, moreover it is common to use such simple models in research literature: they represent building blocks for more complex models used in practice, while are easier to analyse.

Since the proposed objective function is nonsmooth, we couldn't use standard software for the training, and instead implemented a numerical routine from scratch. The relevant optimisation problem turns out to be quasiconvex, it can be solved by the bisection method, with each step requiring a linear feasibility problem to be solved. In~\cite{DANTE} the authors explore the application of quasiconvexity in the case of least squares as well, but their approach is fundamentally different from ours, since our approach uses global characteristics of the functions, rather than local.

We have implemented the algorithm in MATLAB and ran our numerical experiments in tandem alongside a standard implementation with the mean squared loss error function. Our experiments confirm that for a very small training set our max-norm model may perform better than the standard mean least squares formulation, moreover, adding an extra step that removes the outliers may improves the classification further.

The paper is organised as follows. In Section~\ref{sec:NN} we explain a basic model of an artificial neural network and fix the notation that we use in subsequent sections. In Section~\ref{sec:OptMethods} we recap some optimisation background related to quasiconvex functions and the bisection method that are used in Section~\ref{sec:training} to perform the numerical experiments. Theorem~\ref{theorem:main} is the central mathematical result of this paper, connecting neural network models and quasiconvex optimisation. Section~\ref{sec:SizeLim} reports the results of numerical experiments. Finally, Section~\ref{sec:Conc} provides conclusions and future research directions.

\section{Training a Simple Artificial Neural Network}\label{sec:NN}

We consider a basic model of an artificial neural network, and describe a training algorithm that can be used for the model with no hidden layers that is based on a bisection method for quasiconvex programming.

%\subsection{Models of Neural Networks}

A basic model of a neural network consists of several layers of nodes (artificial neurons), connected by directed edges. The network receives a (numerical) signal to its input layer and calculates the output on each subsequent layer using some real valued functions (propagation functions), effectively calculating a composition of these functions as the signal propagates through this network to the output layer. A classic example of an artificial neural network is a handwritten number recognition system, where the input is a grayscale image of a handwritten symbol, and the output is the digit that this handwritten symbol represents \cite{MNIST}. The propagation functions that pass the signal on to the next layer are usually selected from a parametrised family, with parameters (weights) adjusted during the training of the network.

We will use the tuple $x = (x_1,\dots, x_n)\in \R^n$ to represent the input of the neural network (for example, this can be the values of the grayscale pixels in the input image), and $y\in \R^m$ represents the output that encodes the (classification) information extracted by the neural network (for instance, the digit that the handwritten symbol represents). 

\begin{figure}
\centering 
\includegraphics[width=0.5\linewidth]{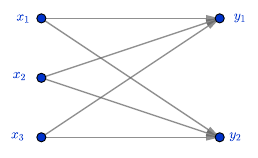}
\caption{A simple neural network without hidden layers}
\label{fig:neural-no-hidden}
% https://www.geogebra.org/m/kx2c4whd
\end{figure}

An elementary example of a neural network consists of one input layer with $n$ nodes and the output layer with $m$ nodes (see Fig.~\ref{fig:neural-no-hidden}).
Usually the propagation function is chosen as a composition of some univariate function $\sigma$ (activation function that plays the role of the sigmoid from models of biological neural networks) with a parametrised affine mapping. The (multilayered) network is then a composition of linear functions with the univariate function $\sigma$. 

In this work we will focus on a simple model with no hidden layers, and hence the input $x$ is converted into the output $y$ by a single composition of an activation function $\sigma$ and some affine (linear) parametrised function, so 
\begin{equation*}
y_i(x) = \sigma\left(\sum_{j=1}^n w_{ij} x_j +w_{i0}\right) \quad \forall i \in \{1,\dots, m\},
\end{equation*}
where $w_{ij}\in \R$  for $i\in \{1,\dots, m\}$, $j\in \{0,\dots n\}$ are the weights. The free parameter $w_{0j}$ is usually called the bias weight, as the model is contextualised as having an additional set of input nodes called biases that always take the value 1, so that the output function is linear on a lifted space with one additional input variable $x_0=1$. 

Most commonly used activation functions are nondecreasing, and in fact it will be convenient for us to assume that $\sigma$ is a continuous strictly increasing function.

We can rewrite the model in a matrix notation as $y(x) =  \varphi(W,x)= \sigma(\bar W x + w_0)$, where 
\begin{equation*}
W = [w_0,\bar W] = \begin{bmatrix}
 w_{10} & w_{11} & \dots & w_{1n} \\
 \vdots & \vdots & \ddots & \vdots \\
 w_{m0} & w_{m1} & \dots & w_{mn}
\end{bmatrix},
\end{equation*}
and we abuse the notation writing $\sigma(p)$ for a componentwise application of $\sigma$ to some vector $p\in \R^d$, so that
\begin{equation*}
\sigma(p) := \begin{bmatrix}\sigma(p_1)\\\sigma(p_2)\\\vdots \\ \sigma(p_d)\end{bmatrix}\quad \text{for}\quad p =  \begin{bmatrix}p_1\\p_2\\\vdots \\ p_d\end{bmatrix}.
\end{equation*}

%\subsection{Loss functions}

The training of a neural network normally consists of minimising a loss function that captures how closely the generated output describes the desired output values. Given a training set $Z = \{(\bar x^1,\bar y^1),(\bar x^2,\bar y^2),\dots,(\bar x^N,\bar y^N)\}$ that consists of pairs of inputs and desired outputs, we would like to choose our parameters $W$ in such a way that the outputs produced by the neural network $y(\bar x_i) = \varphi(W,\bar x^i)=  \sigma(\bar W \bar x^i + w_0)$ are as close as possible to the desired outputs $\bar y^j$ for $j\in \{1,\dots, N\}$. A common loss (error) function used in this context is based on mean least squares formulation,  
\begin{equation*}
L_{1,2}(Z,W) = \sum_{i=1}^N \|\bar y^i-\varphi(W,\bar x^i)\|^2_2,
\end{equation*}
where $\|\cdot\|_2$ is the 2-norm, 
\begin{equation*}
\|p\|^2 = \sum_{i=1}^d p_i^2 \quad \text{for}\quad p = (p_1,\dots, p_d).
\end{equation*}

Our goal is to explore a different choice of the loss function, specifically 
\begin{equation}\label{eq:uniformloss}
L_{\infty,\infty}(Z,W) = \max_{i\in \{1,\dots,N\}} \max_{j\in \{1,\dots, m\}} |\bar y^i_j-\varphi_j(W,\bar x^i)|,
\end{equation}
which is effectively a composition of max-norms. We could have made different choices on both levels, but our interest lies in departing reasonably far from the mean least squares. The nature of this function ensures that we count the contribution from outliers in the dataset, rather than discount them via averaging.

\section{Quasiconvex Functions and the Bisection Algorithm}\label{sec:OptMethods}

The notion of quasiconvexity was originally introduced in the area of financial mathematics~\cite{deFinetti1949}, where the author studied the behaviour of functions with convex sublevel sets, but the term quasiconvexity was introduced much later. 

 Let $D$ be a convex subset of $\R^n$. A function $ f:D\to \R$ is \emph{quasiconvex} if and only if its  sublevel set
	\begin{equation*}
		S_{\alpha}=\{x\in D\, |\, f (x ) \leq \alpha \} 
	\end{equation*}
is  convex for any $\alpha \in \mathbb{R}$. % The set $S_{\alpha}$ is also called an $\alpha$-sublevel set. 
It is not difficult to observe that this definition is equivalent to requiring 
\begin{equation}\label{eq:quasiconvex}
	f (\lambda x + ( 1-\lambda ) y )
	\leq
	\max\{ f ( x ) , f ( y )\} \quad \forall\, x,y\in D , \quad \lambda\in [0,1].
\end{equation}
This characterisation is convenient for proving a number of important properties of quasiconvex functions, for instance, that the supremum of quasiconvex functions is quasiconvex.

A function $f:D\to \R$, where $D$ is a convex subset of $\R^n$,  is called quasiconcave if $-f$ is quasiconvex. Functions that are both quasiconvex and quasiconcave are called quasiaffine (quasilinear). 

A quasiconvex function does not need to be continuous, same applies to quasiconvcave and quasiaffine functions. %Quasiaffine functions are effectively univariate monotone functions
In the case of univariate functions, quasiaffine functions are monotone. 
%in higher dimensions it follows from the definition that their level sets must be half-spaces, and it is then evident that the hyperplanes defining these half-spaces need to be parallel. 
There are many studies devoted to quasiconvex optimisation~\cite{SL,JPCrouzeix1980quasi,DaniilidisHadjisavvasMartinezLegas2002,dutta2005abstract,RubinovSimsek,Rubinov00}. In these studies, the notion of quasiconvexity appears as one of the possible generalisations of convexity.

It is easy to see (e.g. using the characterisation \eqref{eq:quasiconvex}) that if $g:~\R^n\rightarrow\R$ is quasiconvex and $h: \R\rightarrow\R$ is nondecreasing, then the composition $f=h\circ g$ is quasiconvex. It appears that a similar statement is true when $g$ is quasiaffine and $h$ is monotone.

\begin{lemma}[Composition of monotone and quasiaffine functions]\label{lemma:quasiaffine_composition}
Assume that $g:~\R^n\rightarrow\R$ is quasiaffine and $h: \R\rightarrow\R$ is monotone, then the composition $f=h\circ g$ is quasiaffine.
\end{lemma}
{\bf proof}
Assume that $h$ is nondecreasing and $g$ is quasiaffine, then $f=h\circ g$ is quasiconvex. %~\cite{SL}. 
It remains to prove that $-f=-h\circ g$ is quasiconvex. 

Consider an auxiliary function~$\tilde{h}:\R\rightarrow\R$, such that $\tilde{h}(t)=h(-t)$ and therefore $h(t)=\tilde{h}(-t)$. It is clear that if $h$ is nondecreasing then $\tilde{h}$ is nonincreasing and vice versa.  We have
$$-f=-h\circ g=-\tilde{h}\circ(-g).$$
Since $-\tilde{h}$ is nondecreasing and $-g$ is quasiconvex ($g$ is quasiaffine), then $-f$ is quasiconvex and therefore $f$ is quasiaffine. 

It is also clear from the proof that $h$ does not have to be non-decreasing, it is enough for~$h$ to be monotone.
%\end{proof}
\hskip 300pt
$\square$

A standard technique for minimising quasiconvex functions is the bisection method (see \cite[Section~4.2.5]{SL}). If a quasiconvex minimisation problem has an optimal solution and a lower and upper bounds on the optimal value of the objective are known, the method proceeds by bisecting the interval between the lower and upper bounds and solving a feasibility problem to detect whether the sublevel set corresponding to that midpoint is nonempty. If the sublevel set is empty, then the bisector becomes the new lower bound, otherwise the new upper bound is assigned the bisector value, and the process continues.

%\hskip250pt $\square$

\section{Training the Model}\label{sec:training}

We consider a simple network without hidden layers (only input and output layers), furthermore we assume that the output layer consists of a single node.   In this case on the input $x\in \R^n$ the single output neuron produces the output
\begin{equation}\label{input_for_output}
\varphi (W, x) =\sigma\left(\sum_{j=1}^n w_j x_j^i+w_0\right),
\end{equation}
where $\sigma$ is a monotone activation function, and $W$ is reduced to a row vector, so we omit the first index and write $w_j$ for $w_{1j}$ with $j\in \{1,\dots, N\}$. Since the function $\sum_{j=1}^n w_j x_j^i+w_0$ is affine, hence quasiaffine, from Lemma~\ref{lemma:quasiaffine_composition} we conclude that the composition $\varphi (x)=\sigma(\bar w x + w_0)$ is a quasiaffine function. Since changing the sign or adding a constant preserves the quasiaffinity, the loss function \eqref{eq:uniformloss} can be represented as the maximum of quasiaffine functions,
\begin{equation}\label{eq:lossmax}
L_{\infty,\infty} (Z,W) = \max_{i\in \{1,\dots,N\}} \max \{\bar y^i - \varphi (W, \bar x^i),-\bar y^i + \varphi (W, \bar x^i)\}.
\end{equation}
Qiasiaffine functions are quasiconvex, and the maximum of quasiconvex functions is quasiconvex, therefore our loss function is quasiconvex in $W$. We have proved the following result. 

\begin{theorem}\label{theorem:main} In the case of a simple neural network consisting of an input layer and a single output node,  the loss function $L_{\infty,\infty}$ is quasiconvex.
\end{theorem}

\begin{remark} Observe that Theorem~\ref{theorem:main} can be generalised to the case when the output contains more than one node: sandwiching one more layer of maxima in \eqref{eq:lossmax} again results in a quasiconvex function, as the proof is based on the same argument about a maximum of quasiaffine functions. 
\end{remark}

Our next goal is to develop the implementation of the bisection method for our setting. We would like to solve the minimisation problem
\begin{equation*}
\min_{w \in \R^{n+1}}\max_{i\in 1:N} \left|y^i - \sigma\left(\sum_{j=1}^n w_j x_j^i+w_0\right)\right|.
\end{equation*}
Here $ w = (w_0,w_1,\dots, w_n) = (w_0,\bar w)\in \R\times \R^n$ are the weights to be decided, and $Z = \{(\bar x_i, \bar y_i)\}_{i=1}^N$, with  $(\bar x^i,\bar y^i)\in \R^n\times \R$, $i\in \{1,\dots, N\}$ is the training set. 

We know from Theorem~\ref{theorem:main} that the max function is quasiconvex, hence we can apply the bisection method to this minimisation problem. 

It is evident that the objective function is nonnegative, so we can choose $l_0= 0$ as the lower bound for the optimal value. For the upper bound we can substitute any value of the parameter $w$ in the objective, for instance, $w = 0\in \R^{n+1}$, then  
\begin{equation*}
u_0 := \max_{i\in 1:N} \left|\bar y^i- \sigma(0)\right|.
\end{equation*}

We know that the optimal value of the objective function is between $l_0$ and $u_0$. Let $L_1:= \frac{l_0+u_0}{2}$. On each iteration we solve the feasibility problem 
\begin{equation}\label{eq:feasibility}
\max_{i\in 1:N} \left|\bar y^i - \sigma\left(\sum_{j=1}^n w_j \bar x_j^i+w_0\right)\right|\leq L_k.
\end{equation}
If the problem is feasible, we let $l_k = l_{k-1}$, $u_{k} = L_k$. Otherwise we let $l_k = L_{k}$, $u_k = u_{k-1}$. 

We set a threshold $\varepsilon>0$ and continue the process until the gap between the upper and the lower bound is smaller than this value, that is, our stopping criterion is 
$$u_{k_0}-l_{k_0}<\varepsilon.$$ 

This is a simple procedure, which terminates in a finite number of steps. The convergence is linear~\cite{SL} and therefore it is essential to start with the most accurate estimations for $u_0$ and $l_0$ to reduce the number of iterations.

The feasibility problem \eqref{eq:feasibility} that we are required to solve on every iteration 
can be equivalently rewritten as 
\begin{equation*}
\bar y^i - L_k \leq  \sigma\left(\sum_{j=1}^n w_j \bar x_j^i+w_0\right) \leq \bar y^i + L_k \quad \forall i \in \{1,\dots, N\}.
\end{equation*}
Under the assumption that $\sigma$ is strictly increasing and hence has an inverse, this can be rewritten as 
\begin{equation*}
\sigma^{-1}(\bar y^i - L_k) \leq  \sum_{j=1}^n w_j \bar x_j^i+w_0 \leq \sigma^{-1}(\bar y^i + L_k) \quad \forall i \in \{1,\dots, N\}.
\end{equation*}
%Note that if $\varphi$ is nondecreasing, we can replace the inverses with the relevant minima and maxima over the value of the multifunction inverse, as in
%\[
%\inf(\varphi^{-1}(y^i - L_k)) \leq  \sum_{j=1}^n w_j^T x_j^i+w_0 \leq \sup(\varphi^{-1}(y^i + L_k)). 
%\]
Notice that this problem is a linear feasibility problem (in other words, it is a system of linear inequalities with respect to $w$), hence it can be solved with any standard linear programming technique.

\begin{remark}
The bisection method works in the case of non-decreasing functions (for example, classical Rectified Linear Unit (ReLU) activation function is not strictly monotone), since the problem remains quasiconvex. The only difference is that the left- and right-hand side values will correspond to the minima and maxima respectively over the set-valued inverse images of $\sigma$. 
\end{remark}

%\subsection{Activation function is strictly monotone: Leaky Rectified Linear Unit (Leaky ReLU)}

Our assumption of strict monotonicity fares well with the choice of the Leaky Rectified Linear Unit (Leaky ReLU) activation function,
\begin{equation}\label{eq:ReLU}
    \sigma(t)=\alpha t+(1-\alpha)\max\{0,t\},~\alpha>0.
\end{equation}
This is a piecewise linear function with only two pieces, changing the linear slope at the origin: 
\begin{equation*}
    \varphi(t)=
    \left\{
    \begin{array}{rl}
    \alpha t,& t\leq 0;\\
    t,& t>0.
    \end{array}\right.
\end{equation*}
A common choice of the parameter is  $\alpha=0.01$, that is
\begin{equation*}
    \sigma(t)=
   \left\{
    \begin{array}{rl}
    0.01t,& t\leq 0;\\
    t,& t>0.
    \end{array}\right.
\end{equation*}
It is easy to see that  the inverse can be calculated explicitly, 
\begin{equation*}
    \sigma^{-1}(s)={1\over\alpha} s+{\alpha-1\over \alpha}\max\{0,s\}=
    \left\{
    \begin{array}{rl}
    {1\over\alpha} s,& s\leq 0;\\
    s,& s>0;
    \end{array}\right.  
\end{equation*}
%Note that the inverse $\varphi^{-1}$ is closely related to the level sets of~$\varphi$. - obvious
hence the linear feasibility problem \eqref{eq:feasibility} can also be written and coded explicitly.

\section{Numerical Experiments%: classification results when the training set size is limited
}\label{sec:SizeLim}
%\subsection{Motivation}

The goal of our numerical experiments is to test our hypothesis that using a max-type (uniform) loss function \eqref{eq:uniformloss} may be beneficial in training an artificial neural network in the setting of a reliable but small training dataset.

In our numerical experiments, we model this situation by reducing the size of the training set. We use HandOutlines dataset from~\cite{UCRArchive2018}. This dataset contains 2~classes, the corresponding training set contains	1000~recordings (362~recordings in Class~1 and 638~recordings in Class~2) and the corresponding test set contains	370~recordings (133 recordings in Class~1 and 237 recordings in Class~2). Each record contains~2709 floating point values. 
The dataset contains the information about hand outlines of the subjects and their age (image type data). The data set was manually labelled as correct and incorrect hand outlines by three volunteers. If all three volunteers agree that a data point is valid, it is labelled as correct and hence, class 2 contains correctly identified data points whereas class 1 contains incorrectly identified data points. The size of the classes is not equal: Class~2 is almost twice the size of Class~1.
%\todo{It may be worth explaining what these classes are, contextually.}

\subsection{Experiments with the original dataset}

We start with the original dataset (1000~points training set and 370~points test set) and compare the classification results obtained by MATLAB Deep learning toolbox and the results obtained via our uniform approximation algorithm (using bisection method, $\varepsilon=10^{-5}$).

%\todo[inline]{Why are the values of the tolerance are different here and in the later experiments? Is this because of the size of the dataset? Can we provide an explanation for this?}

%\todo[inline]{What is the motivation to use this specific dataset?}

We use the default activation functions for MATLAB Deep learning experiments and Leaky ReLu function with $\alpha=10^{-2}$ for uniform approximation. We report the classification accuracy and provide the confusion matrix. The classification accuracy gives the proportion of points that were assigned the correct class, and the confusion matrix gives more details: the diagonal entries give the number of elements from each of the two classes that were classified correctly, while the off-diagonal elements correspond to the number of misclassified points. In general, the rows of the confusion matrix correspond to the actual class and the columns correspond to the predicted class. For example, the element at the position $(1,2)$ correspond to the number of points from Class~1 assigned to Class~2 (misclassified).

The results of the experiments are shown in Table~\ref{table:original}.
\begin{table}
	\centering
	\caption{Original dataset: classification results}
	\label{table:original}

    \begin{tabular}{|c|c|c|}
    \hline
    %\noalign{\smallskip}
    &Test set &\\
   Method &classification & Confusion matrix\\
   &accuracy &   \\
   \hline
   % Deep learning &    &    \\
    MATLAB toolbox& 89.7\%    & 
        \begin{tabular}{c|c}
           108  & 25 \\
             \hline
           13 & 224
        \end{tabular}
          \\
    \hline
    Uniform approximation  & 60.54\% & \begin{tabular}{c|c}
           77  & 56 \\
             \hline
           90  & 147
        \end{tabular} \\
    \hline
    \end{tabular}

\end{table}

From Table~\ref{table:original} one can see that MATLAB toolbox is much more accurate (almost 90\%), while the uniform approximation is just above~60\%.

Our intention is to demonstrate that uniform approximation is a better tool when the size of the corresponding training set is small. To demonstrate this numerically, we swap training and test sets and now the training set contains 370 points, while the test set contains 1000~points. The results are presented in Table~\ref{table:swapped}.

\begin{table}
    \centering
    \caption{Original dataset: classification results, training and test sets are swapped}
    \label{table:swapped}
    \begin{tabular}{|c|c|c|}
    \hline
    &Test set &\\
   Method &classification & Confusion matrix\\
   &accuracy &   \\
   \hline
   % Deep learning &    &    \\
    MATLAB toolbox& 84.5\%    & 
        \begin{tabular}{c|c}
           222  & 140 \\
             \hline
           15  & 623
        \end{tabular}
          \\
    \hline
    Uniform approximation  & 66.7\% & \begin{tabular}{c|c}
           195 & 167 \\
             \hline
           166  & 472
        \end{tabular} \\
    \hline
    \end{tabular}

\end{table}
One can see from Table~\ref{table:swapped} that the classification accuracy for MATLAB toolbox decreased, which is not surprising, since the size of the training set is reduced and therefore less information is used to train the models. On the other hand, the classification accuracy in the case of uniform approximation has improved. Uniform approximation, due to its nature, treats smaller (under-represented) groups as valid points, while least squares approximation tends to ``average'' and therefore under-represented groups tend to be ``ignored''. This is a great advantage when the under-represented groups are outliers, but in many cases these points are valid data. On the other hand, the presence of ouliers may decrease may decrease the accuracy in the case of uniform approximation. Therefore, our hypothesis is that uniform approximation approach is preferable in the following cases.
\begin{enumerate}
\item Absence (or small number) of outliers.
\item Presence of under-represented groups of  valid  data or uneven distribution of data between the classes (that is, one class is significantly larger than others).
\item Limited size of the available data, where most datapoints are valid and accurate. 
\end{enumerate}
The last case is very common in applications where each datapoint is a result of a very expensive experiment or procedure~\cite{ExpensiveExperiments}.  

Based on our hypothesis, the improvement in the classification accuracy in the case of uniform approximation is due to the fact that many outliers are now removed. 
%\todo{How exactly does the reduction in size affect the accuracy?}. 
Overall, MATLAB toolbox is still more accurate, but this simple experiment encourages us to proceed with the reduction of the training set.
\subsection{Experiments with reduced training sets}
Our next step is to reduce the training set even more. The experimental setup is as follows.
\begin{enumerate}
    \item We use a reduced size test set (the exact size is specified in each experiment) to train the model and the original training set (1000 points) to test (we use this set for testing, since it is a larger set);
    \item The tolerance $\varepsilon$ in the bisection procedure is $\varepsilon=10^{-5}$.
    %\todo{We used the same tolerance earlier}.
\end{enumerate}

Our experiments can be divided into two groups. 
\begin{enumerate}
    \item Equal vs unequal distribution of points between the classes in the training set. In this group of experiments, we are checking if it is harder to create accurate classification rules for unbalanced training sets. 
    \item Random vs non-random selection of reduced size training set. Random selection of training data is a common approach, since it reduces the chance of getting an ``unusual'' piece of data. At the same time, non-random  selection of points (for example, top 10\%) is helpful when one needs to recreate the experiments on the same piece of data.     
\end{enumerate}

%\todo[inline]{We need to explain why we reduced the tolerance here, otherwise it looks weird that the experiments were performed with different tolerancies - can we actually rerun the first batch  with $10^{-5}$?}

\subsubsection{Even number of representatives from each class in the training set}

%\todo[inline]{In this section we need to explain how we chose the points.}

In this experiment, we use 20~points for training: first 10 points from Class~1 and first 10~points from Class~2 (taken from 370~point set, which is the test set for the original dataset). The results are in Table~\ref{table:even}.

\begin{table}
	\centering
    \caption{Reduced dataset: classification results for even number of points from each class in the training set}
	\label{table:even}
    \begin{tabular}{|c|c|c|}
    \hline
    &Test set &\\
   Method &classification & Confusion matrix\\
   &accuracy &   \\
   \hline
   % Deep learning &    &    \\
    MATLAB toolbox& 83.9\%    & 
        \begin{tabular}{c|c}
           218  & 144 \\
             \hline
           17  & 621
        \end{tabular}
          \\
    \hline
    Uniform approximation  & 70.60\% & \begin{tabular}{c|c}
           188  & 174 \\
             \hline
           120  & 518
        \end{tabular} \\
    \hline
    \end{tabular}
\end{table}
The results demonstrate that MATLAB toolbox is still more accurate, but the uniform approximation-based approach is coming closer. 

Our next step is to consider situations, where the size of the training set remains at the same level, but one of the classes is underrepresented.

\subsubsection{Uneven number of representative from each class in the training set}
\paragraph{Training set contains 40 points}
Consider the situation where the training set contains 40~points: 35~points from Class~1 and 5~points from Class~2. The results are in Table~\ref{table:uneven1}.
\begin{table}
    \caption{Reduced dataset: classification results for uneven number of points from each class in the training set: 35~points in Class~1 and 5~points in Class~2}
	\label{table:uneven1}
    \centering
    \begin{tabular}{|c|c|c|}
    \hline
    &Test set &\\
   Method &classification & Confusion matrix\\
   &accuracy &   \\
   \hline
   % Deep learning &    &    \\
    MATLAB toolbox& 74.6\%    & 
        \begin{tabular}{c|c}
             116 & 246  \\
             \hline
             8 & 630 
        \end{tabular}
          \\
    \hline
    Uniform approximation  & 66.5\% & \begin{tabular}{c|c}
             128 & 234 \\
             \hline
             101 & 537 
        \end{tabular} \\
    \hline
    \end{tabular}
\end{table}

The classification accuracy is still higher in the case MATLAB toolbox, but this difference is reducing.

Consider now a symmetric situation where the training set contains 40~points: 5~points from Class~1 and 35~points from Class~2. The results are in Table~\ref{table:uneven2}.
\begin{table}
	 \caption{Reduced dataset: classification results for uneven number of points from each class in the training set: 5~ponts in Class~1 and 35~points in Class~2.}
	\label{table:uneven2}
    \centering
    \begin{tabular}{|c|c|c|}
    \hline
    &Test set &\\
   Method &classification & Confusion matrix\\
   &accuracy &   \\
   \hline
   % Deep learning &    &    \\
    MATLAB toolbox& 64.3\%    & 
        \begin{tabular}{c|c}
            296 &  66 \\
             \hline
            291 & 347
        \end{tabular}
          \\
    \hline
    Uniform approximation  & 69.5\% & \begin{tabular}{c|c}
            193 & 169 \\
             \hline
             136& 502
        \end{tabular} \\
    \hline
    \end{tabular}
\end{table}

In this experiment, the classification accuracy is higher for uniform approximation. 

\paragraph{Training set contains 20 points}

Consider the situation where the training set contains 20~points: 18~points from Class~1 and 2~points from Class~2. The results are in Table~\ref{table:uneven3}.
\begin{table}
	\caption{Reduced dataset: classification results for uneven number of points from each class in the training set: 18~points in Class~1 and 2~points in Class~2}
	\label{table:uneven3}
    \centering
    \begin{tabular}{|c|c|c|}
    \hline
    &Test set &\\
   Method &classification & Confusion matrix\\
   &accuracy &   \\
   \hline
   % Deep learning &    &    \\
    MATLAB toolbox& 68.6\%    & 
        \begin{tabular}{c|c}
            65 & 297\\
             \hline
             17  & 621
        \end{tabular}
          \\
    \hline
    Uniform approximation  & 63.30\% & \begin{tabular}{c|c}
            91 & 271 \\
             \hline
             96& 542
        \end{tabular} \\
    \hline
    \end{tabular}
\end{table}

\begin{table}
	 \caption{Reduced dataset: classification results for uneven number of points from each class in the training set: 2~points in Class~1 and 18~points in Class~2.}
	\label{table:uneven4}
    \centering
    \begin{tabular}{|c|c|c|}
    \hline
    &Test set &\\
   Method &classification & Confusion matrix\\
   &accuracy &   \\
   \hline
   % Deep learning &    &    \\
    MATLAB toolbox& 39.5\%    & 
        \begin{tabular}{c|c}
            338 & 24 \\
             \hline
             582 & 56
        \end{tabular}
          \\
    \hline
    Uniform approximation  & 56.60\% & \begin{tabular}{c|c}
            267 & 95 \\
             \hline
             339& 299
        \end{tabular} \\
    \hline
    \end{tabular}
\end{table}
The classification accuracy is still higher in the case MATLAB toolbox, but this difference is not very larger.

Consider now a symmetric situation where the training set contains 20~points: 2~points from Class~1 and 18~points from Class~2. The results are in Table~\ref{table:uneven4}.

In this experiment, the classification accuracy is low for both approaches, which is not surprising, given the size of the training set and uneven distribution of classes. However, the uniform approximation approach is more accurate.

\paragraph{Conclusions}

Overall, when the size of the training set is reducing, the uniform approximation approach becomes more efficient than MATLAB toolbox (based on MSE). This observation is especially significant when Class~1 is significantly underrepresented.
\subsubsection{Random choice of training set points}

%\todo[inline]{Here we should perhaps run several batches of experiments, as just one successful random example may be attributed to good luck}

%\todo[inline]{Do we actually need the earlier non-random experiments if we now have the random ones?}

In this section we present the numerical results when the training set points were chosen randomly from the original test set (370~points), while the testing was performed on the original training set (1000~points). The total size of the training sets are 100, 50 and 20.  Since the points are chosen randomly, the distribution of the points between classes is also random.

Each experiment was repeated ten times and the reported test set accuracy is the average. This approach is related to the commonly used 10-fold cross-validation approach with some adjustment to our problem. 
\paragraph{Random training set size is 100}

We start with the case where the training set contains 100~points in total. The results are in Table~\ref{table:random100}.
\begin{table}
	\caption{Reduced dataset: classification results for randomly generated training set of 100 points: repeated 10 times, the accuracy is averaged.}
	\label{table:random100}
    \centering
    \begin{tabular}{|c|c|}
    \hline
   Method &Test set classification accuracy\\
   \hline
   % Deep learning &    &    \\
    MATLAB toolbox& 56.94\%     \\
    \hline
    Uniform approximation  & 73.19\%  \\
    \hline
    \end{tabular}
\end{table}

The classification accuracy is  higher for uniform approximation.

\paragraph{Random training set size is 50}
The training set contains 50~points in total. The results are in Table~\ref{table:random50}.
\begin{table}
	\caption{Reduced dataset: classification results for randomly generated training set of 50 points: repeated 10 times, the accuracy is averaged.}
	\label{table:random50}
    \centering
    \begin{tabular}{|c|c|}
    \hline
   Method &Test set classification accuracy\\
   \hline
   % Deep learning &    &    \\
    MATLAB toolbox& 57.08\%     \\
    \hline
    Uniform approximation  & 71.58\%  \\
    \hline
    \end{tabular}
\end{table}

The classification accuracy is  higher for uniform approximation, but the gap is slightly decreasing compared to the experiment with 100~points.

\paragraph{Random training set size is 20}
The training set contains 20~points in total. The results are in Table~\ref{table:random20}.
\begin{table}
	\caption{Reduced dataset: classification results for randomly generated training set of 100 points: repeated 10 times, the accuracy is averaged.}
	\label{table:random20}
    \centering
    \begin{tabular}{|c|c|}
    \hline
   Method &Test set classification accuracy\\
   \hline
   % Deep learning &    &    \\
    MATLAB toolbox& 56.57\%     \\
    \hline
    Uniform approximation  & 69.21\%  \\
    \hline
    \end{tabular}

\end{table}

The classification accuracy is  higher for uniform approximation. The gap between uniform approximation and MATLAB is quite significant.

\paragraph{Why the reduction of the training set leads to the improvement in the classification accuracy in the case of the uniform approximation-based approach?}

One possible explanation is that by removing a significant proportion of points, we also remove all (or almost all) outliers. If this is the case, then the removal of outliers make uniform approximation a better choice. Most outliers are recording or instrumental error and should be removed. At the same time, if recording or instrumental are common, it is unrealistic to avoid them is real-life applications. 

In the next section, we provide the results of numerical experiments where points with high absolute deviation from best uniform approximation are considered as outliers.

\subsubsection{High absolute deviation point removal from the training set}

In this section we assume that points with the highest absolute deviation from best uniform approximation are treated as outliers. The procedure contains two main steps (recall that we use the original test set for training and the training set for testing due to their size).
\begin{enumerate}
    \item{\bf Training set reduction} Find best uniform approximation using the original test set (370~points). Identify the points whose absolute deviation is maximal or close to maximal with a specified tolerance~$\varepsilon$. Refine the training set (370~points) by removing these points (outliers).
    \item{\bf Actual training} Treat the refined  set as the training set and perform test classification on the original training set (1000~points).  
\end{enumerate}

Note that since there are several points with almost the same absolute deviation, it is hard to anticipate the threshold~$\varepsilon$ that gives a certain percentage of data reduction. We consider two cases. In the first case, the threshold~$\varepsilon=10^{-7}$: 46~points out of 370 are removed. In the second case, our goal is to remove (approximately) half of the training set points: 181~points are treated as outliers and the remaining 189~points are used for training.
\paragraph{Training set size is 324}

\begin{table}
	\caption{Reduced dataset: 324 valid points and 46 outliers.}
	\label{table:max324}
    \centering
    \begin{tabular}{|c|c|c|}
    \hline
    &Test set &\\
   Method &classification & Confusion matrix\\
   &accuracy &   \\
   \hline
   % Deep learning &    &    \\
    MATLAB toolbox& 57.3\%    & 
        \begin{tabular}{c|c}
            81 & 281 \\
             \hline
             143 & 495
        \end{tabular}
          \\
    \hline
    Uniform approximation  & 71\% & \begin{tabular}{c|c}
            216 & 146 \\
             \hline
             144 & 494
        \end{tabular} \\
    \hline
    \end{tabular}

\end{table}

We start with the case where 46~points are treated as outliers and therefore they are removed. The results are in Table~\ref{table:max324}.

The classification accuracy is significantly higher for uniform approximation.
\paragraph{Training set size is 189}

\begin{table}
	\caption{Reduced dataset: 181 outliers  and 189 valid points}
	\label{table:max189}
    \centering
    \begin{tabular}{|c|c|c|}
    \hline
    &Test set &\\
   Method &classification & Confusion matrix\\
   &accuracy &   \\
   \hline
   % Deep learning &    &    \\
    MATLAB toolbox& 57.1\%    & 
        \begin{tabular}{c|c}
            92 & 270 \\
             \hline
             159 & 479
        \end{tabular}
          \\
    \hline
    Uniform approximation  & 72.4\% & \begin{tabular}{c|c}
            235 & 127 \\
             \hline
             149 & 489
        \end{tabular} \\
    \hline
    \end{tabular}

\end{table}
In this experiment, 181~points are treated as outliers while 189~points are counted as valid points. The results are in Table~\ref{table:max189}.

The classification accuracy is significantly higher for uniform approximation and the gap is increasing compared to the situation where fewer outliers are removed.
%\subsubsection{Overall conclusions}

\section{Conclusions and Future Research}\label{sec:Conc}

The results of the numerical experiments support our hypothesis that uniform approximation-based approach can be more efficient than mean least squares based approach when the training data is reliable while limited in size. This type of data are common in industrial, medical and research applications, where each datapoint is a result of an expensive experiment, medical procedure or expert evaluation that are not performed routinely. The size of such datasets can be very small, but each recording is carefully performed and therefore is a valid point. Based on our experiments, it may be beneficial to use a uniform approximation based approach, rather than the standard approach for this type of data.

In the absence of an expert opinion it may be hard to distinguish between outliers and valid points that are underrepresented in a given dataset. This is one of the (many) reasons for gender, racial and other kinds of bias present in modern automatic decision making processes  \cite{Gebru} that rely on the standard artificial neural network approach and implicitly assumes Gaussian data distribution. This may lead to automatic discarding of under-represented data points as errors. This problem is not new, but it may be interesting to explore if replacing the mean least squares classifiers with uniform approximation may lead to useful results.

Our work represents an initial step in researching the use of uniform approximation to address problems arising with data classification via standard artificial neural networks when a modest but high quality sample of classified data points is available. The following research themes are of future interest. 
\begin{enumerate}
\item Extend our approach to artificial neural networks with several hidden layers.
\item Study other classification problems with limited or unbalanced training data.
\item Identify the types of activation functions that are efficient for particular datasets, rather than using ``standard'' ReLU activation.
\end{enumerate}

Finally, we would like to emphasise that the essence of deep learning is in approximation and optimisation and therefore there is a need for more robust mathematical models to tackle these problems. On the other hand, some fast  heuristics may be used in the models where the size of complexity makes it hard to apply mathematical optimisation. Generally speaking, this models are not as reliable as those based on mathematical optimisation, but in some cases they are the only way we can handle such  problems.

\section*{Acknowledgement}
This research was supported by the Australian Research Council (ARC), Solving hard Chebyshev approximation problems through nonsmooth analysis (Discovery Project DP180100602).

%\end{acknowledgements}

%\bibliographystyle{spmpsci}
%\bibliography{references}

%\bibitem{SL}
%S.~Boyd and L.~Vandenberghe, \emph{Convex optimization}, Cambridge University
%  Press, New York, NY, USA, 2010.

%\bibitem{JPCrouzeix1980quasi}
%J.~P. Crouzeix, \emph{Conditions for convexity of quasiconvex functions},
%  Mathematics of Operations Research \textbf{5} (1980), no.~1, 120--125.

%\bibitem{DaniilidisHadjisavvasMartinezLegas2002}
%Aris Daniilidis, Nicolas Hadjisavvas, and Juan-Enrique Martinez-Legaz, \emph{An
%  appropriate subdifferential for quasiconvex functions}, SIAM Journal on
%  Optimization \textbf{12} (2002), 407--420.

%\bibitem{deFinetti1949}
%Bruno de~Finetti, \emph{Sulle stratificazioni convesse}, Ann. Mat. Pura Appl.
%  (1949), 173--183.

%\bibitem{dutta2005abstract}
%J~Dutta and AM~Rubinov, \emph{Abstract convexity}, Handbook of generalized
%  convexity and generalized monotonicity \textbf{76} (2005), 293--333.

%\bibitem{Hornik} Kurt Hornik, \emph{Approximation capabilities of multilayer feedforward networks},
%Neural Networks,
%\textbf{4} (1991), no.~2,  251--257.

%\bibitem{RubinovSimsek}
%A.~M. Rubinov and B.~Simsek, \emph{Conjugate quasiconvex nonnegative
%  functions}, Optimization \textbf{35} (1995), no.~1, 1--22.

%\bibitem{Rubinov00}
%Alex~M. Rubinov, \emph{Abstract convexity and global optimization}, Kluwer
%  Academic Publishers, New York, 2000.

%\end{thebibliography}

\end{document}